# A method for estimating forest carbon storage distribution density via artificial intelligence generated content model

Zhenyu Yu[1, 2], Jinnian Wang[1, 2, *]

**Affiliations:**

[1] School of Geography and Remote Sensing, Guangzhou University, Guangzhou, Guangdong, 510006, China; yuzhenyu@gzhu.edu.cn

[2] Innovation Center for Remote Sensing Big Data Intelligent Applications, Guangzhou University, Guangzhou, Guangdong, 510006, China; jnwang@gzhu.edu.cn

Corresponding author: Jinnian Wang (yuzhenyu@gzhu.edu.cn)

**Abstract:** Forest is the most significant land-based carbon storage mechanism. The forest carbon sink can effectively decrease the atmospheric $CO_2$ concentration and mitigate climate change. Remote sensing estimation not only ensures high accuracy of data, but also enables large-scale area observation. Optical images provide the possibility for long-term monitoring, which is a potential issue in the future carbon storage estimation research. We chose Huize County, Qujing City, Yunnan Province, China as the study area, took GF-1 WFV satellite image as the data, introduced the KD-VGG module to extract the initial features, and proposed the improve implicit diffusion model (IIDM). The results showed that: (1) the VGG-19 module after knowledge distillation can realize the initial feature extraction, reduce the inference time and improve the accuracy in the case of reducing the number of model parameters. (2) The Attention + MLP module was added for feature fusion to obtain the relationship between global and local features and realized the restoration of high-fidelity images in the continuous scale range. (3) The IIDM model proposed in this paper had the highest estimation accuracy, with RMSE of 28.68, which was 13.16 higher than that of the regression model, about 31.45%. In the estimation of carbon storage, the generative model can extract deeper features, and its performance was significantly better than other models. It demonstrated the feasibility of artificial intelligence-generated content (AIGC) in the field of quantitative remote sensing and provided valuable insights for the study of carbon neutralization effect. By combining the actual characteristics of the forest, the regional carbon storage estimation with a resolution of 16-meter was utilized to provide a significant theoretical basis for the formulation of forest carbon sink regulation.

**Keywords:** carbon storage; remote sensing; distribution density; diffusion model; AIGC

# 1 Introduction

Due to the escalating threat of global climate change, human activities are inevitably posing unprecedented challenges to the earth's ecosystem (Lenoir et al., 2020). Forests resources management and utilization have become a major global ecosystem issue for researchers and decision-makers (Pecchi et al., 2019). Forest ecosystem not only has a variety of ecological service functions, but also is an important regulator of the global carbon cycle and climate change (Brando et al., 2019). Monitoring and quantifying the temporal and spatial variation characteristics of regional carbon storage is helpful to explore the influencing factors and regulation approaches of forest carbon source/sink function (Nolan et al., 2021; Hua et al., 2022).

Currently, forest carbon storage monitoring is trending towards integrating ground-monitoring sample data and satellite remote-sensing observation data (Hurtt et al. 2019; Sun et al. 2020; Lee et al. 2021; Santoro et al. 2022). While ground monitoring can provide accurate data, it was time-consuming, labor-intensive, and not feasible for large-scale area observations (Xiao et al. 2019; Gao et al. 2023). Remote-sensing inversion method overcame the limitations of sample source monitoring, enabling more efficient and accurate monitoring that was also improving gradually (Long et al. 2020; Gray et al. 2020). The method that used spectral information can deduce the status of vegetation growth and types of vegetation in the ecosystem, allowing for an estimation of carbon storage. Although this method boasted high accuracy, exploring the deep-seated nonlinear relationships was challenging due to image quality, leading to limited improvement in estimation accuracy (Chen et al. 2019). The method of estimating based on structural information can directly measure biomass and carbon storage, but its utility was restricted by the resolution and coverage of remote sensing images (Xiao et al., 2019). The method of estimating based on physical models can estimate carbon storage by creating a carbon cycle model that simulated the carbon sink and cycle processes in the ecosystem. This method can consider the differences and complexities of varying ecosystems, but it required accurate ecological parameters and data support (He et al., 2021). The method of estimating based on machine learning can uncover the deeper relationship between image data and carbon storage and provided fast and efficient estimations (Odebiri et al., 2021; Wang et al., 2022). The integration of sample monitoring data and remote sensing data to develop a universal and high-precision remote sensing monitoring model for regional forest carbon storage had become an urgent task to resolve.

Medium and high resolution (10~30 m) optical data was one of the most promising remote sensing options available as auxiliary datasets. The long lifespan of these satellites made them highly suitable for continuous monitoring of forest dynamics (Puliti et al., 2021). When using multispectral images as data, the calculation of carbon storage mostly involved estimating biomass and stock, with few direct estimation studies of carbon storage from image data. Zhang et al. (2019) used ground observation, MODIS, GLAS, climate, and terrain data in conjunction with the random forest algorithm to produce a 1 km map of aboveground biomass in China with an interpretability of 75% and an RMSE of 45.5 mg/ha. The total change in forest above-ground biomass in a forest area of northern Norway (about 1.4 million hectares) was estimated by Puliti et al. (2021) using National Forest Inventory (NFI),

Sentinel-2, and Landsat data. The aboveground biomass of the southwestern region in the U.S. was estimated from 2000 to 2015 by Chopping et al. (2022) using the Multi-angle Imaging Spectro Radiometer (MISR) with an RMSE of 37.0 Mg/ha. Although the medium and high-resolution remote sensing images contained detailed spatial features and rich surface texture information, the spectral bands were relatively few, and existing carbon storage data products had low resolution. The key to accurately estimating carbon sink was proposing effective algorithms to mine deep-seated features.

With the wide application of deep learning methods, deep-seated features can be excavated to the maximum extent, and the accuracy can be improved to the maximum extent (Xiao et al., 2019; Matinfar et al., 2021; Pham et al., 2023). Most of the current studies used traditional methods such as Ordinary Least Squares (OLS), Random Forest (RF) and Support Vector Regression (SVR), which were simple and efficient, but the estimation accuracy was limited. The application of deep learning method in carbon storage estimation had not been found. The generative model can generate new samples, improve data utilization, enhance the generalization ability of the model, and solve the problem of insufficient or incomplete data, which can be used for the estimation of carbon reserves. The goal of generative model was to model the distribution of real data through learning and processed multimodal data with the method of implicit variable. Generative Adversarial Network (GAN) proposed by Goodflow et al. (2014) was an adversary model composed of generators and discriminators. Through the adversary training of the two networks, GAN can generate high-quality samples. However, GAN might have problems such as instability and mode collapse in the training process, which required careful adjustment and attention to the training parameters. The Variational Autoencoders (VAE) proposed by Kingma & Welling (2013) made the generated samples had better continuity and controllability by learning the continuous potential space, but the performance of VAE in generating high-quality realistic samples was worse than that of GAN, which might be because the potential space of VAE may be fuzzy. The diffusion model was first proposed in 2015 (Sohl-Dickstein et al., 2015). Its purpose was to eliminate the Gaussian noise continuously applied to the training image, which can be regarded as a series of de-noising self encoders. It used a variant called latent diffusion model (LDM) to train the self encoder to convert images into low dimensional potential space. The diffusion model had good sample generation quality and high flexibility. At the same time, the training process was relatively stable and was not prone to mode collapse. It made up for the shortcomings of GAN and took into account the advantages of VAE and GAN. DDPM (Ho et al., 2020), DDIM (Song et al., 2020), stable diffusion (Rombach et al., 2022) and other models proposed later applied the idea of diffusion model to image generation and refreshed state-of-the-art (SOTA) records. Nowadays, they had been widely used in the field of image translation. Thus, we applied this structure to estimate carbon storage of remote sensing images with improved accuracy.

To sum up, we transformed the problem of carbon storage estimation into the task of image translation, utilized the current remarkable implicit diffusion model as the fundamental network structure, and proposed an improved algorithm. Taking GF-1 WFV image as the main data, combined with the actual characteristics of Huize County, it realized the estimation of regional high-resolution carbon storage distribution density, and provided an important theoretical basis for the formulation of forest carbon sink regulation.

In summary, our main contributions include:

(1) The VGG-19 module after knowledge distillation is added to the diffusion model to realize the initial feature extraction and constructed the encoder-decoder structure.

(2) Attention + MLP was added for feature fusion to obtain the relationship between local and global features and realized the restoration of high-fidelity images in the continuous scale range.

(3) IIDM reduced the inference time and improved the accuracy while reducing the number of model parameters. The experimental results showed that AIGC was feasible in remote sensing.

## 2 Material and methods

### 2.1 Study area

Huize County is subordinate to Qujing City, Yunnan Province, China. Yunnan Province is located in Southwest China. Qujing City is the second largest city in Yunnan Province. Huize County is located in the northeast of Yunnan Province and the northwest of Qujing City. It is located at the junction of Yunnan Province, Sichuan Province and Guizhou Province. The total land area is 5,889 km$^2$, as shown in Figure 1. The terrain of Huize County is characterized by a ladder-like decline from the west to the east, with higher peaks in the west and lower valleys in the east. Huize County is characterized by a typical temperate plateau monsoon climate. The county of Huize has notable climate features, spanning from a subtropical zone in the south to a cold temperate zone in the north. The altitude of Huize County varies significantly, with the highest peak reaching a staggering 4,017 m, surpassing the rest of Qujing City. The lowest altitude in Qujing City is 695 m, marking the lowest point in the city. According to the Forest Management Inventory data, the forest land area in Huize County is about 3,080.53 km$^2$, and the arbor forest land is about 2,538.20 km$^2$, accounting for 82.39%. The Huize County boasts an abundant supply of forest resources, presenting a diverse range of samples for research; however, the intricate physical and geographical landscape presents significant challenges to the study.

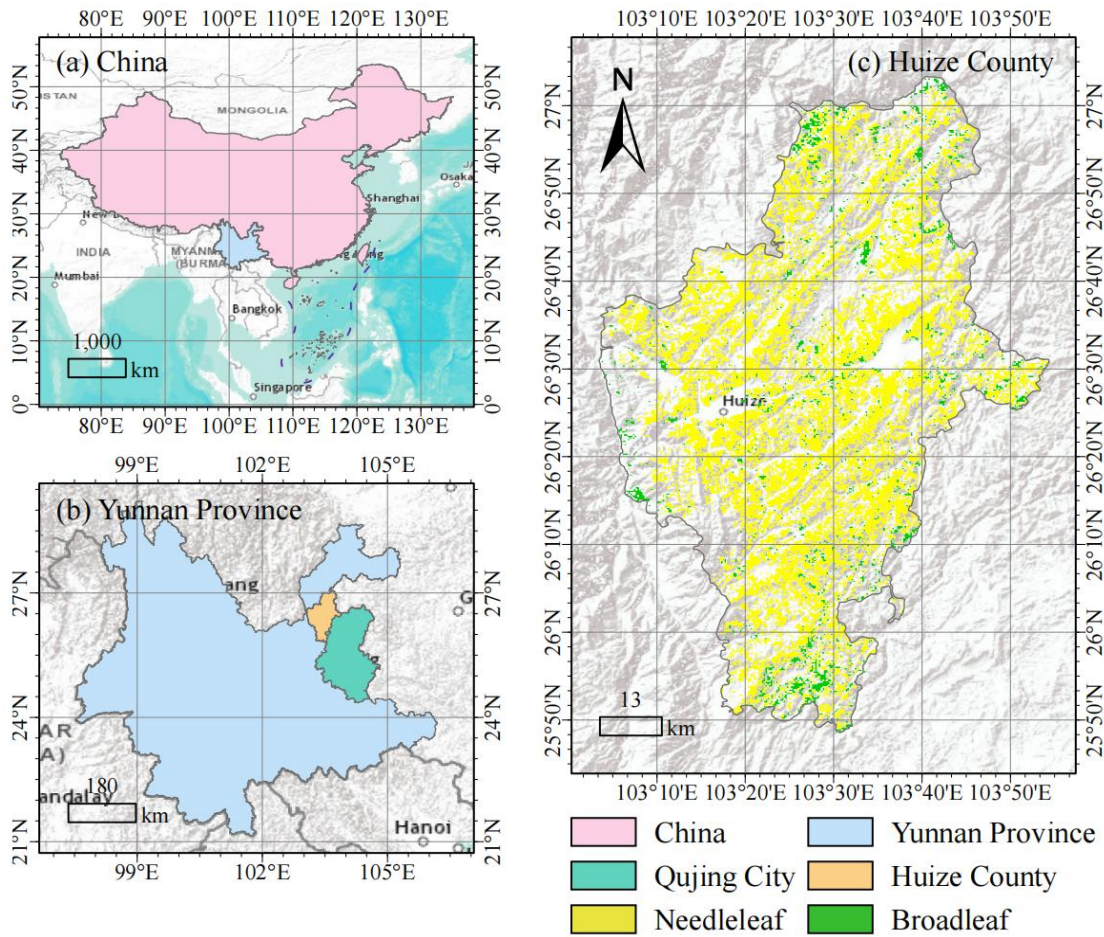

**Figure 1** Study area. Notes: (a) China, (b) Yunnan Province, and (c) Huize County.

**2.2 Data sources**

**Survey data.** The study utilized the data from the Forest Management Inventory which was obtained from the Huize County Forestry and Grass Bureau and spanning from 2020. This data referred to data obtained through systematic and comprehensive investigation and monitoring of forest resources. It can be used to understand and assess the status, change trend, distribution, and ecological environment of forest resources, providing scientific basis for forestry management, protection, and sustainable utilization. The data included more than 70 attributes such as forest area and type, which can objectively represent the overview of forest resources in the study area and assist in forest resource estimation.

**Satellite imagery.** The study mainly used GF-1 WFV image data. GF-1 satellite was a high-resolution remote sensing satellite successfully launched by China in 2013. It was widely used in agriculture, forestry, environmental monitoring, urban planning, resource investigation, disaster monitoring and other fields. The spatial resolution of GF-1 WFV data was 16 m. In order to ensure the time consistency, the imaging time for participating in the model training was August 27, 2020, including two images, which realized the full coverage of the study area. The data information of the images is shown in Table A1.

**Elevation data.** The ALOS PALSAR DEM was selected for the digital elevation model (DEM) data, with a spatial resolution of 12.5 m, and the four-view image achieved full coverage of the study area. The data information is shown in Table A1. ALOS PALSAR is a satellite equipped with a Synthetic Aperture Radar (SAR) sensor that scans the surface through a radar beam and measures the topographic altitude and surface characteristics. From 2006 to 2011, PALSAR's L-band SAR provided detailed and all-weather interference measurements.

**Canopy height.** The canopy height data we selected is the 10-meter vegetation canopy height data set of ETH Global Sentinel-2 (Lang et al., 2022). By fusing GEDI with Sentinel-2, Lang et al. (2022) developed a probabilistic deep learning model to retrieve canopy height from Sentinel-2 images anywhere on Earth. The global wall-to-wall map is based on Sentinel-2 images taken between May and September 2020. We have extracted the data in Huize County.

**2.3 Methods**

**2.3.1 Preprocessing**

(1) Calculate carbon storage

The survey data did not include carbon storage, but per-hectare accumulation data was contained. The calculation of forest carbon storage is shown in Eq. (1). The default values of conversion factors for climate change projections had been used for the Intergovernmental Panel on Climate Change (IPCC). The volume expansion coefficient $\delta$ was generally taken as 1.90, and the bulk density coefficient, i.e., the dry weight coefficient, $\rho$, was generally taken as 0.45 ~ 0.50 t/m³. We took 0.5 t/m³ in this study. The carbon content rate $\gamma$ is generally taken as 0.5, $V$ is the accumulated volume (m³), and $C$ is the carbon storage (t or Mg). The accumulated volume is calculated as shown in Eq. (2), where $V_{ha}$ is the volume per hectare (m³/ha) and $Area$ is the plaque area (ha).

$$C = 2.439 \times (\delta \times \rho \times \gamma \times V) \tag{1}$$

$$V = V_{ha} \times Area \tag{2}$$

(2) Calculate distribution density

We used the plaque of survey data as a unit area, normalized the canopy height within the area as a weight, and calculated the carbon storage density as shown in Eq. (3). Where, $i$ was the area number, $C$ was the carbon storage, $W$ was the weight, $CD$ was the carbon storage of each pixel, that was, the carbon storage density.

$$CD_i = C_i \times W_i \qquad (3)$$

(3) Identify forest/non-forest areas

On the basis of previous research, we selected F-Pix2Pix proposed by Yu et al. (2023) to extract the forest area, regarded it as a mask, and binarized the extracted image. The forest area was designated as 255, and the non-forest area was set to 0.

**2.3.2 KD-VGG module**

**(1) Source model**

The input before the $relu4\_1$ layer of VGG-19 was regarded as the source model ($ENC$), and the feature knowledge was extracted from the $ENC$ to a smaller target model ($enc$), whose structure followed the source model, but a smaller channel length was used at each layer. Feature knowledge would be extracted from the $reluN\_1(N=1,2,3,4)$ layer of $ENC$ to the corresponding layer in $enc$, which was called $reluN\_1_e$ layer. $\mathbf{F}_{N,k}$ was the feature of the image $I_k$ extracted at the $reluN\_1$ layer of $ENC$, $\mathbf{F}_{N,k} \in C_N \times H_{N,k}W_{N,k}$, and $\mathbf{F}^e_{N,k}$ was the feature extracted at the corresponding $reluN\_1_e$ layer of $enc$, $\mathbf{F}^e_{N,k} \in C^e_N \times H_{N,k}W_{N,k}(C^e_N \ll C_N)$.

**(2) Global eigenbases**

We introduced a global and image independent eigenbases $\mathbf{W}_{N,g} \in \mathsf{R}^{C^e_N \times C_N}$, that was, we established a unique $C^e_N$ dimensional space, which can well capture the global features of the image, as shown in Eq. (4).

$$\max_{\mathbf{W}_{N,g}\mathbf{W}_{N,g}^{\mathrm{T}}=\mathbf{I}} \frac{1}{M} \sum_{k=1}^{M} \mathrm{tr}(\mathbf{W}_{N,g}\overline{\mathbf{F}}_{N,k}\overline{\mathbf{F}}_{N,k}^{\mathrm{T}}\mathbf{W}_{N,g}^{\mathrm{T}}) \qquad (1)$$

Where $M$ was the number of images, and the solution of $\mathbf{W}_{N,g}$ is the eigenbases of $\frac{1}{M}\sum_{k=1}^{M}\overline{\mathbf{F}}_{N,k}\overline{\mathbf{F}}_{N,k}^{\mathrm{T}}$. Take mini-batch gradient descent to minimize the loss $\sum_{I_k \in \beta_t} \mathrm{tr}(\mathbf{W}_{N,g}\overline{\mathbf{F}}_{N,k}\overline{\mathbf{F}}_{N,k}^{\mathrm{T}}\mathbf{W}_{N,g}^{\mathrm{T}})/|\beta_t|$, where $\beta_t$ was a batch of sampled images at the $t^{\mathrm{th}}$ iteration of gradient descent. The minus of a trace as a loss function made the gradient descent process unstable. Such losses had no lower bound, so gradient descent algorithms tended to minimize losses, but ignored the constraint of $\mathbf{W}_{N,g}\mathbf{W}_{N,g}^{\mathrm{T}}=\mathbf{I}$. This was unsolvable in the current situation, and we used Eq. (4) rewriting as Eq. (5) to solve and optimize the problem of $\mathbf{W}_{N,g}$. Among them, Eq. (5) was approximately solvable in the case of small batch gradient descent, where $\mathbf{W}_{N,g}^{\mathrm{T}}\mathbf{W}_{N,g}\overline{\mathbf{F}}_{N,k}$ was the reconstruction characteristic of $\overline{\mathbf{F}}_{N,k}$ according to the $\mathbf{W}_{N,g}\overline{\mathbf{F}}_{N,k}$ map.

$$\max_{\mathbf{W}_{N,g}\mathbf{W}_{N,g}^{\mathrm{T}}=\mathbf{I}} \frac{1}{M} \sum_{k=1}^{M} \| \mathbf{W}_{N,g}^{\mathrm{T}}\mathbf{W}_{N,g}\overline{\mathbf{F}}_{N,k} - \overline{\mathbf{F}}_{N,k} \|_2^2 \qquad (2)$$

In summary, to derive $\mathbf{W}_{N,g}(N=1,2,3,4)$, we used a small batch of gradient descent to simultaneously solve $\mathbf{W}_{N,g}$ in Eq. (5). In the $t^{\mathrm{th}}$ iteration of gradient descent, we sampled the batch $\beta_t$ of the following minimization problem, as shown in Eq. (6), calculated the gradient of the target to update $\mathbf{W}_{N,g}$. The batch size we used was 8, and trained on $\mathbf{W}_{N,g}$ for 5 epochs. Using the global eigenbases $\mathbf{W}_{N,g}(N=1,2,3,4)$, we can extract information from the *reluN*_1 layer of the source model *ENC* to the *reluN*_$1_e$ layer of the target model *enc*, as shown in Figure 2 (a).

$$\min_{\substack{\mathbf{W}_{N,g}\mathbf{W}_{N,g}^{\mathrm{T}}=\mathbf{I} \\ N\in\{1,2,3,4\}}} \frac{1}{|\beta_t|}\sum_{N=1}^{4}\sum_{I_k\in\beta_t} \|\mathbf{W}_{N,g}^{\mathrm{T}}\mathbf{W}_{N,g}\overline{\mathbf{F}}_{N,k} - \overline{\mathbf{F}}_{N,k}\|_2^2 \qquad (3)$$

**(3) Blockwise PCA knowledge distillation**

In order to realize the feature transformation of course-to-fine in the distillation model, in addition to extracting the input information to the encoder $enc$, we also needed to implement the paired decoder $dec$. The distillation method we used was Principal Components Analysis (PCA).

The encoder $enc$ was divided into a series of blocks $\{enc_1, enc_2, enc_3, enc_4\}$, in which the output $enc_N$ was the $reluN\_1_e$ layer, and the decoder $dec$ was divided into a group of blocks $\{dec_4, dec_3, dec_2, dec_1\}$, in which the output $dec_N$ was the $relu(N-1)\_1_d$ layer of reproduction characteristics $relu(N-1)\_1_e$. That was, the decoder took the $relu4\_1_e$ features from $dec$ as input to progressively reproduce the $relu4\_1_e, relu3\_1_e, relu2\_1_e, relu1\_1_e$ features and the reconstructed image. To implement our $enc-dec$ model, we trained each pair of $enc_N$ and $dec_N$ with other pairs by minimizing the distillation loss $\mathrm{L}_{enc}^{N}$ of the encoder and the implementation loss $\mathrm{L}_{dec}^{N}$ of the decoder. These four pairs were trained in order from $N=1$ to $N=4$, as shown in Figure 2 (b), with $enc_2$ and $dec_2$ as examples.

In the distillation of encoder, given the image $I_k$, we wanted to train the encoder block $enc_N$ to make its decentralized output $\overline{\mathbf{F}}_{N,k}^{e}$ close to the feature $\mathbf{W}_{N,g}\overline{\mathbf{F}}_{N,k}$. It was derived from the $\overline{\mathbf{F}}_{N,k}$ map of the global eigenbases $\mathbf{W}_{N,g}$ from the decentralized output of $ENC_N$, with reconstruction loss as shown in Eq. (7).

$$\mathsf{L}_{enc}^{N}(I_k) = \| \mathbf{W}_{N,g}^{\mathrm{T}} \overline{\mathbf{F}}_{N,k}^{e} - \overline{\mathbf{F}}_{N,k} \|_2^2 \tag{4}$$

In the decoder implementation, given an image $I_k$, we wanted to approximate the output $\mathbf{F}_{N-1,k}^{d}$ of the $dec_N$ to the input image $I_k$ in order to reproduce the input $\mathbf{F}_{N-1,k}^{e}$ of the $enc_N$ and the reconstructed image from the $dec_1$. Overall, we had minimized the $I_{k_{rec}}$ formed by three regular terms, as shown in Eq. (8).

$$\mathsf{L}_{dec}^{N}(I_k) = \| \mathbf{F}_{N-1,k}^{d} - \mathbf{F}_{N-1,k}^{e} \|_2^2 + \| I_{k_{rec}} - I_k \|_2^2 + \| \mathbf{F}_{N,k_{rec}} - \mathbf{F}_{N,k} \|_2^2 \tag{5}$$

The third one was to facilitate image reconstruction based on perceived loss. Note that when $N=1$, there was no first term for feature reconstruction. In summary, in training $enc_N$ and $dec_N$, we had addressed the following optimization issues:

$$\min_{enc_N, dec_N} \mathsf{L}_{enc}^{N}(I_k) + \mathsf{L}_{dec}^{N}(I_k) \tag{6}$$

**(4) Reducing channel lengths**

According to the empirical rules in PCA dimensionality reduction, the most important information of the channel length $C_N^e$ of the target model $enc$ should be retained. The target layer $reluN\_1_e$ of $C_N^e$ needed to retain the variance information of more than 85 % of the source layer $reluN\_1$.

For each image $I_k$ in our dataset, we calculated the covariance of its features (called $\mathbf{F}_{N,k}$) extracted in the $reluN\_1$ layer. Let $\sigma_{N,k}^{j}$ be the $j^{th}$ largest eigenvalue of the covariance associated with the $j^{th}$ principal eigenvector $\mathbf{e}_{N,k}^{j}$. The $j^{th}$ explanatory variance (EV) $\sigma_{N,k}^{j} / \sum_{a=1}^{C_N} \sigma_{N,k}^{a}$ reflected the portion of the feature variance captured by $\mathbf{e}_{N,k}^{j}$, while the cumulative EV (CEV) $\sum_{\beta=1}^{C_N'} \sigma_{N,k}^{\beta} / \sum_{a=1}^{C_N} \sigma_{N,k}^{a}$ reflected the feature variance captured by the top $C_N'$ feature vector. We used the mean cumulative explanatory variance (mCEV) to determine the value of $C_N^e$. Mean CEV (mCEV) was the average of the CEV values of all images, as shown in Eq. (10).

$$\mathrm{mCEV}(C_N^{'}) = \frac{1}{M} \sum_{k=1}^{M} \frac{\sum_{\beta=1}^{C_N^{'}} \sigma_{N,k}^{\beta}}{\sum_{\alpha=1}^{C_N} \sigma_{N,k}^{\alpha}} = \sum_{\beta=1}^{C_N^{'}} \mathrm{mEV}(\beta) \qquad (7)$$

Where $M$ was the number of images in the dataset, and $\mathrm{mEV}(\beta) = \frac{1}{M} \sum_{k=1}^{M} (\sigma_{N,k}^{\beta} / \sum_{\alpha=1}^{C_N} \sigma_{N,k}^{\alpha})$ was the mean $\beta$-th EV。 We were looking for $C_N^e$ that meet $\mathrm{mCEV}(C_N^e) \approx 85\%$.

In the original 512 (256, 128, 64) dimensional space of the $relu4\_1 (relu3\_1, relu2\_1, relu1\_1)$ layer, 85% of the variance information can be explained by an average of 64 (58, 20, 5) principal components. In other words, in the four layers, a small part of the original channel length was required to retain more than 85% of the variance information, that was, 12.5% (22.7%, 15.6%, 7.8%).

Chiu & Gurari (2022) found that setting $C_1^e$ to 5 and $C_2^e$ to 20 impeded distillation from the $relu2\_1$ layer to the $relu2\_1_e$ layer. We suspected that was due to the low reduction rate of 7.8% in the $relu1\_1_e$ layer. We found through experiments that doubling the reduction rate and setting $C_1^e$ to 10 can overcome this issue. For our final model, we set the four channel lengths $(C_1^e, C_2^e, C_3^e, C_4^e)$ to $(10, 20, 58, 64)$.

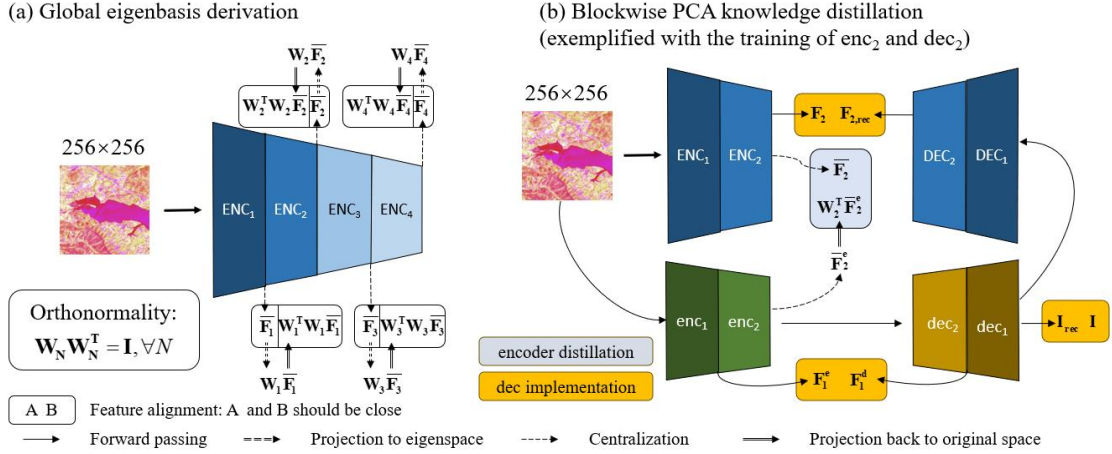

**Figure 2** PCA-based knowledge distillation structure for VGG. Notes: PCA-based knowledge distillation for photorealistic style transfer consists of two steps: (a) global eigenbasis ( $W_N, N = 1, 2, 3, 4$ ) derivation and (b) blockwise PCA knowledge distillation.

### 2.3.3 Denoising model

Conditional network used a Convolutional Neural Networks (CNN) to extract conditional features. We used KD-VGG to establish initial features $\mathbf{f}^{(0)}$. VGG had a large number of parameters. Taking the optimized KD-VGG module as a feature extractor not only deepened the number of network layers and enabled deeper features to be extracted, but also significantly reduced the number of parameters, improving network performance without incurring unnecessary overhead. Finally, concatenated $\mathbf{f}^{(0)}$ and $\mathbf{y}_t$, and sent them to U-Net for encoding. At the same time, it was sent to the CNN to be down sampled to $\mathbf{f}^{(i)}$, as shown in Eq. (11).

$$\mathbf{f}^{(i)} = \mathrm{Conv}(\mathbf{f}^{(i-1)}) \tag{1}$$

Where, the serial number $i$ represented the output of different layers, and $t$ was the time. Unlike GAN-based methods that relied on additional prior knowledge, this conditional network only provided encoded features and sent them to U-Net without requiring additional prior knowledge to model potential representations.

This study was an estimation task, and the scale of the input and output image had not changed. The scaling factor modulation (i.e., scaling factor $s$) of the original model had been modified. By setting this factor to 1, the zoom control for that particular part is removed. On the basis of the original model (Gao et al., 2023), attention was added to fuse the features of $\mathbf{f}^{(i)}$ and $\mathbf{u}^{(i)}$, and then input into the implicit representation module.

**2.3.4 Implicit neural representation**

Currently popular algorithms were usually affected by complex cascading or two-stage training strategies. We found that using implicit neutral representation to learn the representation of continuous images helped to improve the accuracy of estimation. We inserted several coordinates based MLPs into the up-sampling of U-Net architecture to parameterize the implicit neural representation, which can restore high fidelity images in a continuous scale range. Assuming that the continuous coordinates of the multi-resolution feature were referred to as $c = \{c^{(1)},...,c^{(i)},...,c^{(N)}\}$, entered the current feature and the corresponding coordinates to calculate the target feature. The specific process was as follows:

$$\mathbf{u}_{\mathbf{up}}^{(i)} = D_i(\hat{\mathbf{h}}^{(i+1)}, c^{(i)} - \hat{c}^{(i+1)}) \tag{1}$$

$D$ was a two-layer MLP, $\hat{\mathbf{h}}^{(i+1)}$ and $\hat{c}^{(i+1)}$ were interpolated by calculating the closest Euclidean distance between $\mathbf{h}^{(i+1)}$ and $c^{(i+1)}$ at a depth of $i+1$ that achieved up-sampling.

**2.3.5 Improved implicit diffusion model (IIDM)**

Implicit diffusion model (IDM) integrated the benefits of diffusion models and implicit neural representations to create an end-to-end model structure that enabled image-to-image transformation and estimation of carbon stocks. The model structure is shown in Figure 3. IDM used Recurrent Neural Network (RNN) to build a time-dependent network. Each unit contained two modules: denoising model and implicit representation. Parameters were shared each unit.

IDM models through variational inference, mainly because the diffusion model was also a latent variable model. Compared with the implicit variable model such as Variational Auto-encoders (VAE), the implicit variables of the diffusion model were the same dimension as the original data, and the reasoning process (i.e., diffusion process) was often fixed.

**Diffusion principle.** The diffusion model included two processes: diffusion process (i.e., forward process) and reverse process. The forward process was also called diffusion process. Both the forward and reverse process were a parameterized Markov chain, in which the reverse process can be used to generate data. Here we modelled and solved it through variational inference.

**Diffusion process.** The diffusion process refers to the process of gradually adding Gaussian noise to the data until the data became random noise. For the original data $\mathbf{x}_0 \pi \sim q(\mathbf{x}_0)$, each step of the diffusion process with a total of $T$ steps was to add Gaussian noise to the data $\mathbf{x}_{t-1}$ obtained in the previous step as follows:

$$q(\mathbf{x}_t \mid \mathbf{x}_{t-1}) = \mathsf{N}\,(\mathbf{x}_t; \sqrt{1-\beta_t}\,\mathbf{x}_{t-1}, \beta_t \mathbf{I}) \tag{1}$$

Where $\{\beta_t\}_{t=1}^{T}$ was the variance used for each step, which was between 0 and 1. For the diffusion model, we often called the variance of different steps as variance schedule or noise schedule. Usually, the later step would adopt a larger variance, that was $\beta_1 < \beta_2 < ... < \beta_T$. Under a designed variance schedule, if the number of diffusion steps was large enough, the final result $\mathbf{x}_T$ would completely lose the original data and become a random noise. Each step of the diffusion process generated a noisy data $\mathbf{x}_t$. The whole diffusion process was also a Markov chain:

$$q(\mathbf{x}_{1:T} \mid \mathbf{x}_0) = \prod_{t=1}^{T} q(\mathbf{x}_t \mid \mathbf{x}_{t-1}) \tag{2}$$

**Reverse process.** The diffusion process was to noise data, and the reverse process was to remove noise. If we knew the real distribution $q(\mathbf{x}_{t-1} \mid \mathbf{x}_t)$ of each step in the reverse process, then started from a random noise $\mathbf{x}_T \sim \mathsf{N}\,(0, \mathbf{I})$, the gradual denoising can generate a real sample, so the reverse process was also the process of generating data.

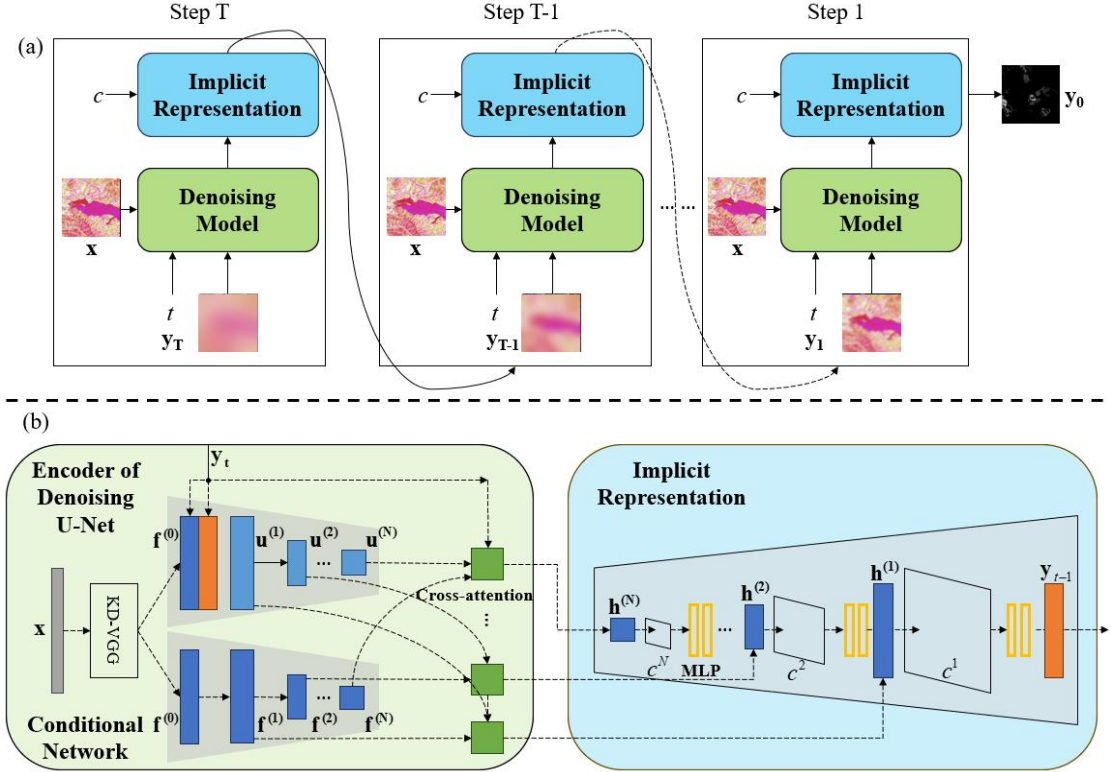

**Figure 3** Improved implicit diffusion model architecture. Notes: (a) Overall process of the inference. It is a reverse process. (b) Detailed illustration of a denoising step, including denoising model and implicit representation. The denoising model includes KD-VGG feature extractor, encoder of denoising U-Net, and conditional network. MLP represents multi-layer perceptron.

### 2.3.6 Optimization

The loss function of $L_1$ was also referred to as the Mean Absolute Error (MAE) loss. $L_1$ loss had a stable gradient for any input value, which would not lead to the gradient explosion problem, and had a more robust solution. However, the process was not smooth at the zero point, it did not exhibit differentiability at this point, and the convergence rate was slow. In general, $L_1$ regularization resulted in sparse features and assigned a weight of 0 to most redundant ones, thus performing feature selection. Additionally, $L_1$ loss was not susceptible to noise and therefore it was more appropriate for regression problems. The $L_1$ distance was presented in Eq. (5). $x$ was the ground truth, and $G(x)$ was the generated images.

$$L_1(G) = \mathbb{E}_{x \sim P_{data}(x)} \left[ \| x - G(x) \| \right] \tag{1}$$

**2.3.7 Evaluation metrics**

Mean Absolute Error (MAE) was the average of the absolute error between the predicted and the real value (Eq. (16)). Mean Squared Error (MSE) was the average of the absolute square error between the predicted and the true value (Eq. (13)). Root Mean Squared Error (RMSE) was the square root of MSE (Eq. (14)). Peak Signal-to-noise Ratio (PSNR) was a full reference image quality evaluation index (Eq. (12)), where $MAX$ was the maximum pixel value of the image. Structure Similarity Index Metrics (SSIM) was also a full reference image quality evaluation index, which measured image similarity in terms of brightness, contrast and structure (Eq. (15)). Where, $\hat{y}$ was the predicted value, $y$ was the real value, $\mu_{\hat{y}}$ and $\mu_y$ was the mean of $\hat{y}$ and $y$ respectively, $\sigma_{\hat{y}}^2$ and $\sigma_y^2$ was the variance of $\hat{y}$ and $y$ respectively, and $\sigma_{\hat{y}y}$ was the covariance of $\hat{y}$ and $y$, $c_1 = (k_1 L)^2$, $c_2 = (k_2 L)^2$, which was generally taken as $c_3 = c_2 / 2$.

$$MAE = \frac{1}{n} \sum_{i=1}^{n} |\hat{y}_i - y_i| \tag{1}$$

$$MSE = \frac{1}{n} \sum_{i=1}^{n} (\hat{y}_i - y_i)^2 \tag{2}$$

$$RMSE = \sqrt{\frac{1}{n}\sum_{i=1}^{n}(\hat{y}_i - y_i)^2} \quad (3)$$

$$PSNR = 10 \cdot \log_{10}\left(\frac{MAX^2}{MSE}\right) \quad (4)$$

$$SSIM(\hat{y}, y) = \frac{(2\mu_{\hat{y}}\mu_y + c_1)(2\sigma_{\hat{y}y} + c_2)}{(\mu_{\hat{y}}^2 + \mu_y^2 + c_1)(\sigma_{\hat{y}}^2 + \sigma_y^2 + c_2)} \quad (5)$$

## 3 Results

### 3.1 Inference time

We used images with a resolution of $256 \times 256$, and the test results using seven models are shown in Table 1. The models included VGG19, KD-VGG19, SD (Stable Diffusion), DDPM, DDIM, Ours-VGG and Ours-KD-VGG, among which DDPM, DDIM and SD are diffusion models. The diffusion model had a large number of parameters (greater than 450M). Further adding modules to achieve higher accuracy would result in excessive system overhead and reduced feasibility. Therefore, it was worth discussing whether such addition was justified.

The parameter of VGG-19 was 78.14M. After reducing the channel dimensionality, the parameter was 290K and the model was compressed by approximately 279.07 times. The KD-VGG-19 model preserved most of VGG19's feature extraction capabilities while reducing the model's size. As a result, VGG19 can now be considered a lightweight feature extraction module that offered better feature extraction capabilities, making it useful for downstream tasks.

The VGG module was added to the feature extraction section of IDM. KD-VGG preserved the model's robust feature extraction ability and improved estimation accuracy while decreasing the inference time when model parameters were scaled down. The inference time of the VGG module was 2.04 times higher compared to that of stable diffusion, while that of KD-VGG module was 2.39 times higher. KD-VGG module required only 21.56% of the parameters as compared to stable diffusion. Ours-KD-VGG showed an improved performance by being 0.07s faster than DDIM.

**Table 1** Size and inference time of different models. Notes: DDPM = Denoising Diffusion Probabilistic Models, and DDIM = Denoising Diffusion Implicit Models. The experiment was implemented in 24GB NVIDIA GeForce RTX 4090 GPU.

| Model | Size | Inference time (Second/Image) |
|---|---|---|
| VGG19 | 78.14M | 0.08 |
| KD-VGG19 | 290K≈0.28M | 0.06 |
| Stable diffusion | 2034.05M | 1.65 |
| DDPM | 461.01M | 1.61 |
| DDIM | 455.77M | 0.76 |
| Ours-VGG | 516.35M | 0.81 |
| Ours-KD-VGG | 438.49M | 0.69 |

**3.2 Ablation study**

Four modules were chosen for the ablation experiment: Mask, VGG, KD-VGG, and Attention + MLP. Table 2 shows the experimental results. Mask was utilized to filter vegetation and non-vegetation areas, and its effect was most evident. This module can be viewed as the initial data preprocessing stage. We employed the VGG and KD-VGG modules for the initial feature extraction. We found the performance of the KD-VGG and VGG models to be comparable. However, the VGG model turned out to be slightly superior to the KD-VGG model, which indicated that the knowledge distillation method was effective. Most of the important parameters of the model were extracted, but the student model still lacked some critical parameters, leaving room for improvement. By introducing attention and MLP modules, we achieved a performance boost in our model. However, the inclusion of the attention mechanism led to an increase in the time of influence. As evident from the results, the accuracy significantly improved after increasing the module. Specifically, in the experiment with mask and KD-VGG (No.10 & 11), the MAE decreased by 1.90. This demonstrated that utilizing the joint KD-VGG and Attention + MLP modules can optimize model accuracy. Therefore, it can be inferred that the strategy of sacrificing time for accuracy was worthwhile.

**Table 2** Compare of modules, including Mask, VGG, KD-VGG, and Attention + MLP. Notes: Bold is the best, and underline is the second.

| No. | Mask | VGG | KD-VGG | Attention + MLP | MAE | RMSE | SSMI | PSNR |
|---|---|---|---|---|---|---|---|---|
| 1 | | | | | 28.1184 | 39.7301 | 0.5653 | 16.1484 |
| 2 | | ✓ | | | 26.7018 | 36.3281 | 0.5765 | 16.9260 |
| 3 | | ✓ | | ✓ | 26.5816 | 35.7425 | 0.6023 | 17.0671 |
| 4 | | | ✓ | | 26.7120 | 36.0418 | 0.5740 | 16.9947 |
| 5 | | | ✓ | ✓ | 26.7353 | 35.8997 | 0.5858 | 17.0290 |
| 6 | | | | ✓ | 27.8921 | 39.1927 | 0.5546 | 16.2667 |
| 7 | ✓ | | | | 23.8728 | 28.7722 | 0.6863 | 18.9513 |
| 8 | ✓ | ✓ | | | 23.8422 | 28.6723 | 0.6853 | 18.9816 |
| 9 | ✓ | ✓ | | ✓ | <u>23.7409</u> | **28.4431** | **0.7055** | **19.0513** |
| 10 | ✓ | | ✓ | | 25.5869 | 28.6836 | 0.6949 | 18.9781 |
| 11 | ✓ | | ✓ | ✓ | **23.6913** | <u>28.5909</u> | <u>0.6956</u> | <u>19.0063</u> |
| 12 | ✓ | | | ✓ | 24.2106 | 28.6897 | 0.6874 | 18.9763 |

**3.3 Comparison experiments**

We chose a total of seven comparison models, including the most commonly used regression models in remote sensing estimation (OLS, RF and SVR), and the generative models of deep learning (VAE, GAN, and IIDM proposed in this paper). The results of the accuracy evaluation are presented in Table 3. OLS was the simplest and least expensive algorithm in all models. RF and SVR were the most widely used machine learning algorithms for estimation. In this task, the accuracy of the three algorithms was the same, and the estimation accuracy was poor in all models. The effect of deep learning algorithm in estimation was significantly improved, the performance of generative model was significantly better than VAE, and diffusion was better than GAN. Among all the models, VGG-IDM performed best, followed by KD-VGG-IDM, but the accuracy difference between the two was small. Considering the parameters and influence time, KD-VGG-IDM was more universal.

The spatial distribution results are shown in Figure 4. The spatial distributions of OLS, RF, and SVR were quite different from the ground truth. The three algorithms were basically higher than the ground truth. RF was the highest, followed by OLS. Although the results of high value areas were relatively close, the overall difference was very large. Both RF and SVR had more unstable point noise, indicating that the estimation results were unstable. VAE, GAN and IIDM were relatively close to the ground truth. VAE had relatively more low value regions, while GAN was generally higher. The details of IIDM were more similar, but the performance of extreme value regions was not as good as GAN.

**Table 3** Compare of carbon stock estimation. Notes: The following results have all added masks. OLS = Ordinary Least Squares, RF = Random Forest, SVR = Support Vector Regress, VAE = Variational Autoencoder, and GAN = Generative Adversarial Network. Bold is the best, and underline is the second.

| Models | MAE | RMSE | SSMI | PSNR |
|---|---|---|---|---|
| OLS | 29.3093 | 41.8396 | 0.5638 | 15.6991 |
| RF | 29.1595 | 41.7539 | 0.5371 | 15.7169 |
| SVR | 26.3446 | 44.4820 | 0.5031 | 15.1671 |
| VAE | 26.9627 | 36.8072 | 0.5516 | 16.8121 |
| GAN | 25.7670 | 35.2341 | 0.6089 | 17.1915 |
| Ours-KD-VGG | **23.6913** | <u>28.6836</u> | <u>0.6956</u> | <u>18.9781</u> |
| Ours-VGG | <u>23.8422</u> | **28.4431** | **0.7055** | **19.0513** |

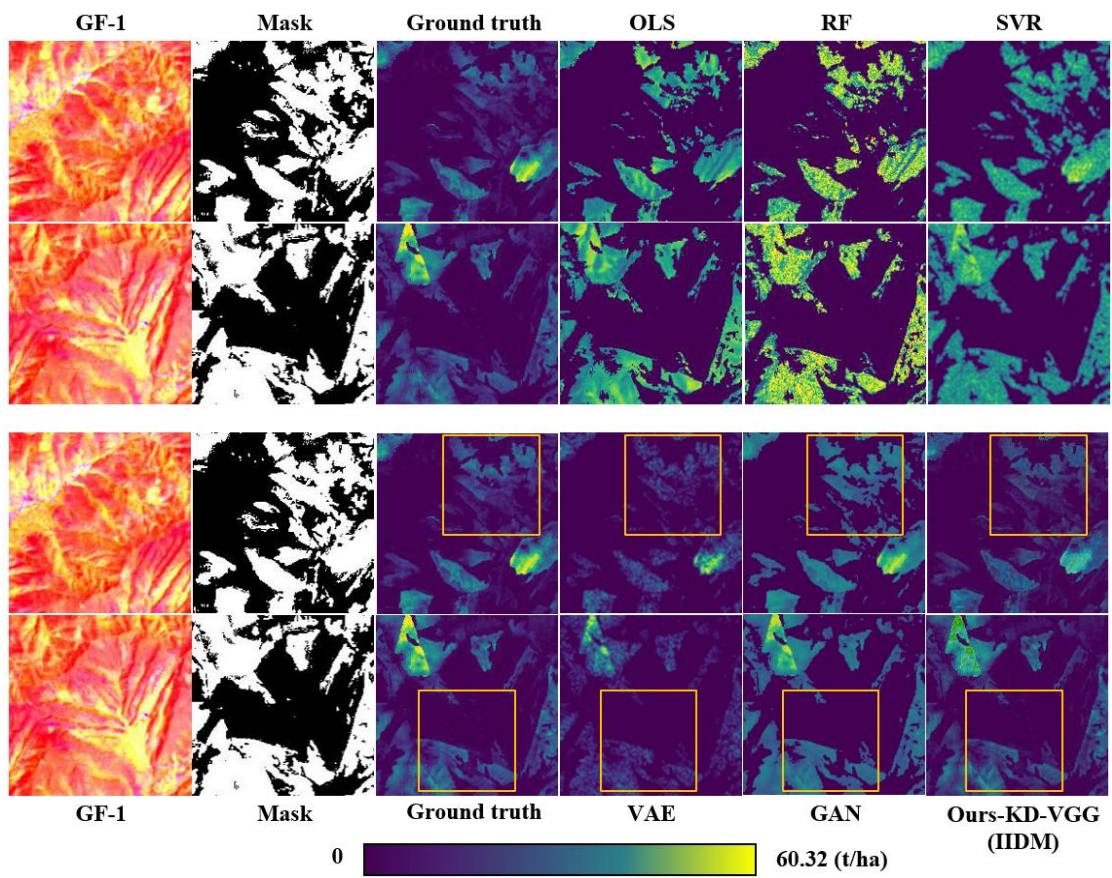

**Figure 4** Carbon stock estimation results.

## 4 Discussion

**4.1 Estimation methods**

The estimation was a regression problem. OLS was the simplest, most intuitive, easy to understand and implement, and provided a clear estimation of the parameters, which made the results more explanatory. However, OLS was sensitive to outliers and outliers, which could lead to unstable results. It performed poorly on complex nonlinear problems and was prone to under-fitting (Venter et al., 2020). RF was not easily affected by overfitting and can effectively handle a large number of features and high-dimensional data. Due to the integration strategy of multiple decision trees, outliers had less impact on RF. Compared with OLS, RF was difficult to provide clear parameter estimation and had poor interpretability (Sakamoto, 2020). SVR could handle complex data relationships through various kernel functions. SVR had strong robustness to deal with a small number of outliers. SVR was based on the theory of support vector machine and had a solid mathematical foundation. However, the choice of parameters in SVR (such as kernel function, penalty parameter, etc.) had a great impact on the performance of the model, which had to be carefully adjusted. As with RF, the interpretability of SVR results was poor (Zhang et al., 2022).

GAN can generate realistic samples, but the training process was unstable due to the simultaneous training of the discriminator and the generator. The main optimization goal of GAN was to make the images realistic, which resulted in a lack of image diversity. The generation of GAN was implicit, implemented by the network, did not follow the probability distribution, and was not interpretable (Li et al., 2021).

VAE can learn the potential variable representation of data, can learn the probability distribution, and had strong interpretability. It introduced the randomness of potential variables, so the generated samples had a certain diversity. However, since the generation process of VAE involved the sampling of potential variables, the reconstruction results might produce some fuzziness (Du et al., 2021). Diffusion models can generate high quality realistic samples. The generation process was controllable. The variety and quality of the sample generation can be controlled by different steps. It was suitable for multimodal data. Based on the theoretical foundation of Markov Chain Monte Carlo (MCMC), it had strong mathematical interpretability. However, training diffusion models was expensive and required multi-step sampling. Compared with other generation models, the training and inference process might be more time-consuming (Yang et al., 2022). This study revealed that AIGC possesses remarkable feasibility in the field of quantitative research in remote sensing.

**4.2 Estimation results**

Huize County boasted abundant forest resources, serving as a solid foundation for research. However, the substantial altitude variation created a more complex natural geographical environment, which posed significant challenges to the precise calculation of carbon storage. Currently, the investigation of carbon storage estimation utilizing satellite imagery was primarily segregated into two main categories: optical imagery and lidar. The accuracy of lidar estimation was superior to that of multispectral images, however, achieving large-scale estimation was relatively challenging, and data acquisition was a major difficulty. Optical imagery addressed this deficiency. GF-1 had the advantage of providing high temporal and spatial resolution. Thus, achieving high precision estimation of optical images could be an upcoming challenge in the research of carbon storage estimation.

In the estimation of multispectral carbon storage, prior research typically employed machine learning models, such as OLS, RF, and SVR (Zhang et al., 2022; Zhang et al., 2019; Cho et al., 2021). Because of the limitations of models, it was possible to extract only linear and shallow nonlinear features. However, the relationship between spectral & textural features and biomass & stock volume & carbon storage was not a simple linear relationship. We proposed the IIDM algorithm to extract deeper features. This model was built on the diffusion model of deep learning and incorporated the addition of the VGG module after knowledge distillation to extract initial features, the utilization of Attention + MLP for feature fusion, and the incorporation of Mask to filter out non-vegetation areas.

This model improved the estimation accuracy (RMSE=28.68) to a level comparable to lidar (RMSE=25.64) (Cao et al., 2016) and outperformed the estimation accuracy of multi-source and multi-temporal coarse resolution (1 km) imagery (RMSE $\approx$ 30) (Chen et al., 2023). Cao et al. (2016) and Chen et al. (2023) conducted biomass estimates that were strongly correlated with carbon storage and were found to be comparable. The results demonstrated that AIGC is appropriate in remote sensing parameter inversion.

**5 Conclusion**

We have selected Huize County as the study are, utilized GF-1 WFV images as data, and employed the generative diffusion model of deep learning. After knowledge distillation, we incorporated the VGG module to extract initial features, then applied Attention + MLP for feature fusion and included a mask to filter out non-vegetation areas. Finally, the IIDM model was proposed which served as the foundation for precise estimation of regional carbon sinks. The main conclusions are:

(1) The VGG-19 module after knowledge distillation was added to the diffusion model to realize the initial feature extraction. The encoder and decoder structure were constructed in the denoising model. IIDM reduced the influence time and improved the accuracy while reducing the number of model parameters.

(2) The cross-attention + MLP module was added for feature fusion to obtain the relationship between global and local features and realized the restoration of high-fidelity images in the continuous scale range. Adding attention enhanced the accuracy and also amplified the inference time; however, the overall time was still shorter compared to other diffusion models.

(3) In the estimation of carbon storage, the generative model can extract deeper features, and its performance was significantly better than other models, which also showed the feasibility of AIGC in remote sensing. The IIDM proposed in this paper has the highest estimation accuracy, with a RMSE of 28.68, which was 13.16% higher than the regression model, approximately.

**Acknowledgements**

We are grateful to AI Earth for providing the support (https://engine-aiearth.aliyun.com). This research was funded by the National Key R&D Program of China [2021YFE0117300].

**Author contributions**

Jinnian Wang supervised and organized the project. Zhenyu Yu developed the code and wrote the manuscript. All authors revised the manuscript.

**Competing interests**

The authors declare no competing interests.